\documentclass{article}
\usepackage[square,numbers,sort]{natbib}
\usepackage{natbib}
\usepackage[final]{nips_2017}
\usepackage[utf8]{inputenc} 
\usepackage[T1]{fontenc}    
\usepackage{hyperref}       
\usepackage{url}            
\usepackage{booktabs}       
\usepackage{amsfonts}       
\usepackage{nicefrac}       
\usepackage{microtype}      
\usepackage{graphicx}

\title{House Price Prediction using Satellite Imagery}

\author{
  Sina Jandaghi Semnani\\
  Stanford University\\
  \texttt{sinaj@stanford.edu}\\
  \And
  Hoormazd Rezaei\\
  Stanford University\\
  \texttt{hoormazd@stanford.edu}\\
}

\begin{document}

\maketitle

\begin{abstract}
In this paper we show how using satellite images can improve the accuracy of housing price estimation models. Using Los Angeles County's property assessment dataset, by transferring learning from an Inception-v3 model pretrained on ImageNet, we could achieve an improvement of ~10\% in R-squared score compared to two baseline models that only use non-image features of the house.
\end{abstract}

\section{Introduction}	

High-resolution satellite images from most of the Earth's surface have been available for over a decade, and in recent years advanced deep learning methods have been applied to them. Mnih et. al. started this field with a classifier that detects roads \cite{mnih}. More recently, deep learning models and satellite imagery have been used for more complicated tasks such as crop yield prediction in the U.S. \cite{cropyield} and poverty prediction in Africa \cite{jean}. \par

We believe there are still many unexplored questions and potential applications for this field of study.
One such problem is using satellite images to enhance existing house price prediction models, which historically only used hand-selected features (e.g. number of rooms or floor area). We show that a vision-based model can learn important relevant features from the neighborhood of a house (e.g. distance from green fields or highways) by processing a single satellite image, and then use those features to improve the estimation accuracy.

Although high-resolution satellite images contain an abundance of information that might be correlated with house prices, such data are highly unstructured and thus challenging to extract meaningful insights from. Although deep learning models such as convolutional neural networks (CNN) could in principle be trained to directly estimate prices from satellite imagery, the scarcity of training data makes the application of these techniques challenging. Therefore, we use transfer learning techniques to overcome this problem in that we will use knowledge gained while solving one problem and apply it to our different but related problem. We will start with a CNN model that has been trained on ImageNet \cite{imagenet}, a large image classification data set, to identify low-level image features such as edges and corners that are common to many vision tasks. Next, we will build on the knowledge gained from this image classification task and fine-tune the model on a new task.

We combine a dataset of house prices (from LA County's property assessment dataset \footnote{\href{https://data.lacounty.gov/Parcel-/Assessor-Parcels-Data-2017/vak5-2hqh}{https://data.lacounty.gov/Parcel-/Assessor-Parcels-Data-2017/vak5-2hqh}}) and a dataset of satellite images (collected via Google Static Maps API \footnote{\href{https://developers.google.com/maps/documentation/static-maps/intro}{https://developers.google.com/maps/documentation/static-maps/intro}}) to accomplish this objective.
The input to our model is a $640\times640$ image (with the house located at the center) and a feature vector (number of bedrooms, etc.). The output is a single number, the value of the house in dollars. We feed the image to the pretrained CNN whose last two layers have been deleted after pretraining, and get a feature vector for the image. We then combine the two feature vectors using several fully-connected layers and get the output.

Finally, we evaluate our estimation performance using $R^2$ and mean-squared error (MSE) metrics. The results will be compared to two baseline models that only use hand-selected features.

The inspiration for this project came from Zillow’s Home Value Prediction competition \footnote{\href{https://www.kaggle.com/c/zillow-prize-1}{https://www.kaggle.com/c/zillow-prize-1}}, which is challenging the data science community to help push the accuracy of the house price estimations further by utilizing any additional source of information \footnote{We even tried reducing Zillow's model's estimation error (which is what the competition is about) using satellite images, but it proved to be a challenging task. Zillow's model already performs exceptionally well, and improving a blackbox model without having access to its details is difficult. We are not sure if Zillow's model uses any vision-based features or not.}. 

\section{Related work}
Traditionally, house price prediction models used hand-selected features (e.g. number of rooms or floor area). Feed-forward neural networks and various decision-tree-based models \cite{dectree} outperformed other approaches. However, as \cite{Chopra} mentions, the parameters that determine the price of a house are twofold: one that is intrinsic to the house and one that indicates the "desirability" of the house. Traditional models can only learn the former, but the latter usually depends on the location of the house \cite{Kockelman}. In particular, social and economic metrics such as crime rate, pollution levels and distance from important locations (e.g. train station, hospitals) can affect the housing price \cite{Bency}. Therefore, a model that is able to capture the resulting spatial correlations can potentially outperform other models. Spatial auto-regressive (SAR) models are one approach to capturing the existing spatial correlation in housing prices. These models rely on a spatial contiguity matrix that is usually hand-designed with the help of a domain expert, but as \cite{Chopra} showed, can also be learned by algorithms. Non of these models use vision-related features. More recently, \cite{Bency} proposed a model that uses deep neural networks to extract information from satellite images, combined them with an SAR model and other house features. This approach resulted in a $57\%$ reduction in RMSE compared to a model with SAR but without satellite images.

It is worth noting that these papers use different datasets which in most cases are gathered by the authors of the paper. This makes it difficult to compare their results in a completely fair manner \footnote{We also use a dataset that we collected.} However, it can be seen that the focus of the research has been moving towards models that can capture features from outside the building itself, which has generally improved the accuracy of estimators.

\section{Dataset and Features}

Our dataset consists of two main parts: 1. real estate properties, and 2. high-resolution satellite images. Details of each part are explained in this section.

\subsection{Real Estate Properties}
We collected properties data from LA County's property assessment data which is available online\footnote{\href{https://data.lacounty.gov/Parcel-/Assessor-Parcels-Data-2017/vak5-2hqh}{https://data.lacounty.gov/Parcel-/Assessor-Parcels-Data-2017/vak5-2hqh}}. This dataset consists of more than 2 million real-estate properties in LA county assessed between 2006 and 2017. Each entry had 50 features, floor area, number of bedrooms, year built, use type, latitude and longitude to name a few. Preprocessing was done on the data, details of which will be discussed later in this paper.

\subsection{High-Resolution Satellite Images}
\begin{figure}
\includegraphics[width=0.5\textwidth]{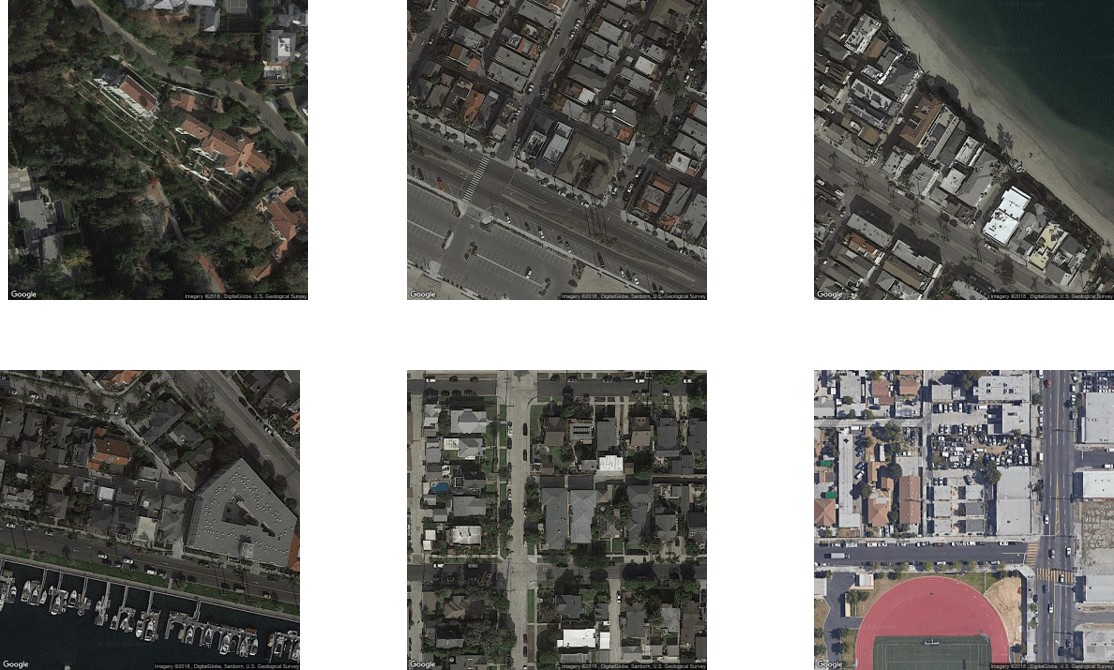}
\centering
\caption{Examples of satellite images. Each image is $640\times640$ pixels.}
\label{fig:sample}
\end{figure}

To better predict the price error, we combined our main dataset with satellite images of houses, collected via Google Static Maps API \footnote{\href{https://developers.google.com/maps/documentation/static-maps/intro}{https://developers.google.com/maps/documentation/static-maps/intro}} using the longitude and latitude of each house. This API gives access to different zoom levels between 1 (the Earth) and 24 (the most detailed). In this study, we set the zoom level to 19, and retrieved images with size 640x640 which is the highest resolution available for the free tier users\footnote{We planned to try different zoom levels, but the limit of 25,000 images per day for the free tier made image gathering very time-consuming}. Figure \ref{fig:sample} shows a few examples. We believe this zoom level provides a suitable amount of information about the neighborhood while showing the shape of the building itself.

\subsection{Data Preprocessing}

Before beginning analysis, the LA County dataset was preprocessed. Eight columns contained unrelated information such as assessor's ID and parcel ID, which were removed. One redundant column was also removed. With each property, came three price values: total land value, personal property value, and total value. In addition to residential properties, the data included rows for non-residential properties as well as empty lots. Since our aim is to study housing price estimations, we focused our analysis on entries for which personal property value was non-zero, and used this column as our labels. Consequently, non-residential rows were removed. Finally, we ended up with 40 features per property, among which 23 were categorical parameters (e.g. use type). We replaced categories with integer values. Each column in the dataset was then normalized to have zero mean and standard deviation of $1$. Table \ref{table:dataset} shows a summary of the size of our dataset.

\begin{table}{}
\centering
\begin{tabular}{|c|c|c|c|}
\hline
Train Set Size & Validation Set Size&Test Set Size&Number of Features \\
\hline
48,548&2,698&2,698&40\\
\hline
\end{tabular}
\vspace{6px}
\caption{Train/validation/test size of the dataset}
\label{table:dataset}
\end{table}

\section{Methods}

\subsection{Baseline Models}

We started our analysis by building two baseline models: decision-tree-based models and neural networks. For the former, we analyzed multiple tree-based estimators. Extra tree regressor \cite{extratrees} demonstrated the best performance and was chosen as our baseline. Moreover, we trained a neural network on our features dataset. Performance of these models are provided in the Experiments and Results section.

\subsection{Inception-v3}

Since we do not have enough data to train vision-based convolutional neural networks, we use pretrained models to encode images. We use Inception-v3 model \cite{inception}, trained on ImageNet, to construct 2048-dimensional vectors for our images. The process is as follows. First, JPEG satellite images are converted into RGB matrices, resized to $299\times299\times3$ and normalized to have values in $[0, 1]$ to match the expected input of Inception-v3. Feeding the data into the model, we then save resulting vectors in a binary file to be used later in our network. This time-consuming process generated around 8 gigabytes of processed data. We do this in order to make the training process faster and be able to iterate more quickly in hyperparameter tuning part of the project.

\subsection{Cost Function}
We use mean-squared error (MSE) as our cost function:
$
J= \frac{1}{m} \sum_{i=1}^{m} L(\hat{y}^{(i)}, y^{(i)})=\frac{1}{m} \sum_{i=1}^{m}(\hat{y}^{(i)}-y^{(i)})^2
$. 
Note that we report our results using $R^2$ (not MSE), but the two metrics are nicely related in a way that minimizing MSE is equivalent to maximizing $R^2$:
$
R^2=1-\frac{MSE}{S}
$
where $S$ is the variance of the test set's labels.

\section{Experiments and Results}

\subsection{Experiments}
All numerical results are provided in table \ref{table:results}.
We have studied to see if images alone are able to make estimations about prices. Encodings are fed into fully-connected neural network layers and trained to minimize mean squared error (MSE) loss. 

From the results we got, it is apparent that by themselves, images are not able to predict prices. This can be attributed to the difficulty of estimations of house size from images. It is not possible for a model to decide on the size of a house by just looking at a satellite picture, as is may include several other buildings too.

Next step was to train models on both features and images to see if combining the two can improve our estimations. We feed our two feature vectors to two separate dense networks and we concatenate the results in the final layer. We will describe the details of this model in the Network Architecture section.

\begin{table}{}
\centering
\begin{tabular}{|c|c|c|c|c|c|c|}
\hline
Model& Train $R^2$ & Dev $R^2$& Test $R^2$& Train MSE& Dev MSE& Test MSE\\
\hline
Extra Tree (baseline) &
0.99
&0.86&
0.71&
0.000&
0.112&
0.002\\
\hline
Neural Net F (baseline) &
0.96
& 0.84
&0.85
&  0.037
&  0.136
&  0.001\\
\hline
Neural Net I &
$0.000$&-0.0001&-0.038&
 1.062
 & 0.859
 & 0.010\\
\hline
Neural Net F+I &
  0.98
&  0.94
&  0.93
&  0.011
&  0.044
&  0.001
\\
\hline
\end{tabular}
 \vspace{0.6em}
\caption{Summary of results. MSE (mean-square error) is calculated using normalized labels. F is features, I is satellite images.}
\label{table:results}
\end{table}
\subsection{Interpretation of the Results}
\begin{figure}
\includegraphics[width=0.6\textwidth]{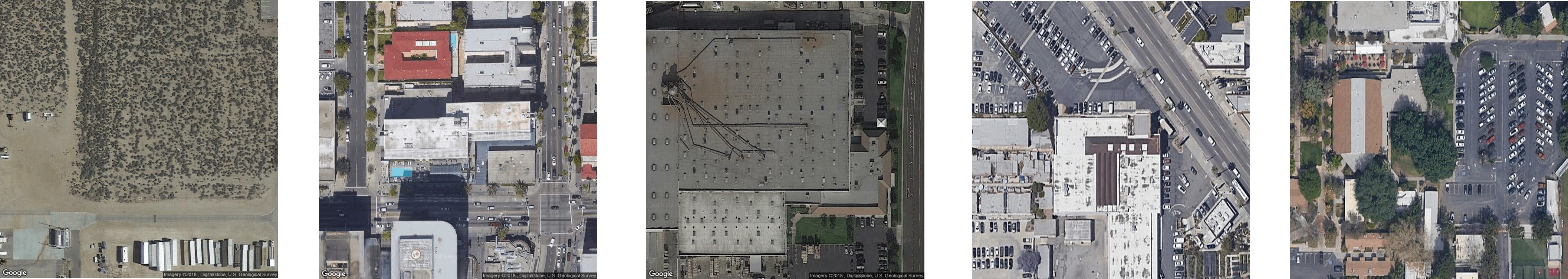}
\centering
\caption{Baselines underestimated the price of these houses and F+I corrected them.}
\label{fig:top}
\end{figure}

\begin{figure}
\includegraphics[width=0.6\textwidth]{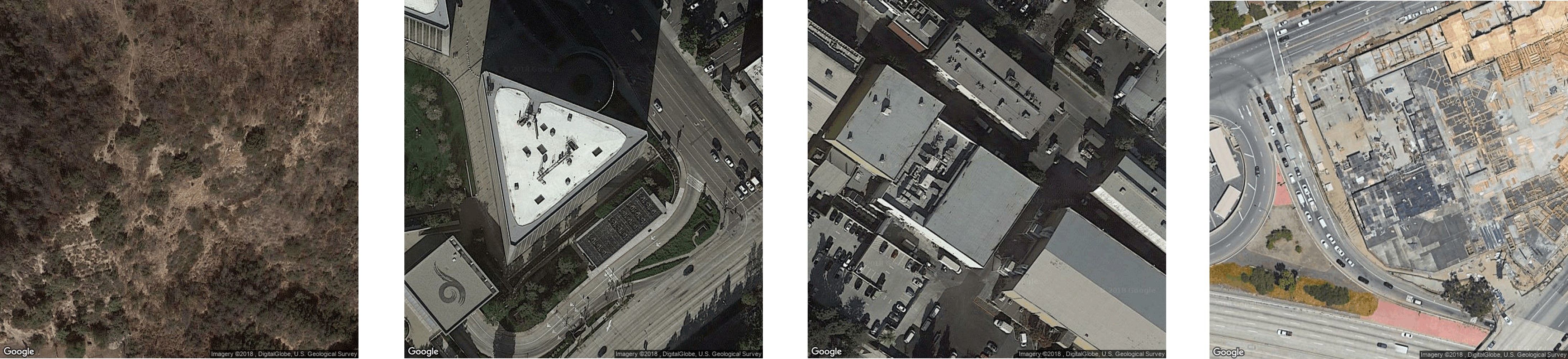}
\centering
\caption{Baselines overestimated the price of these houses and F+I corrected them.}
\label{fig:bottom}
\end{figure}

We see that the F+I model outperforms all other models by at least $10\%$ in $R^2$ metric. Intuitively, the cost of a house is approximately floor area $\times$ price per square foot. While features provide the model with the first factor, images help it improve its estimation for the second. We ranked data points based on how much F+I improves estimations compared to F model. Top 5 positive (F+I corrected underestimation) and top 4 negative improvements (F+I improved overestimation) are shown in figures \ref{fig:top} and \ref{fig:bottom} respectively.

An interesting observation is that our dataset -despite all the preprocessing that we have done- contained a few empty lands with no buildings\footnote{Perhaps due to changes that occurred between the time LA County has assessed properties and the time Google has captured their images}. The fact that these data points are present in figures \ref{fig:top} and \ref{fig:bottom} shows that our model is more robust to noise in the data.
After evaluating figure \ref{fig:top}, we think it shows that our model realizes that cars are an indicator of higher prices (since it shows higher accessibility or low distance from city center), so it adds to the estimated price.

\subsection{Network Architecture}
\begin{figure}
\includegraphics[width=0.6\textwidth]{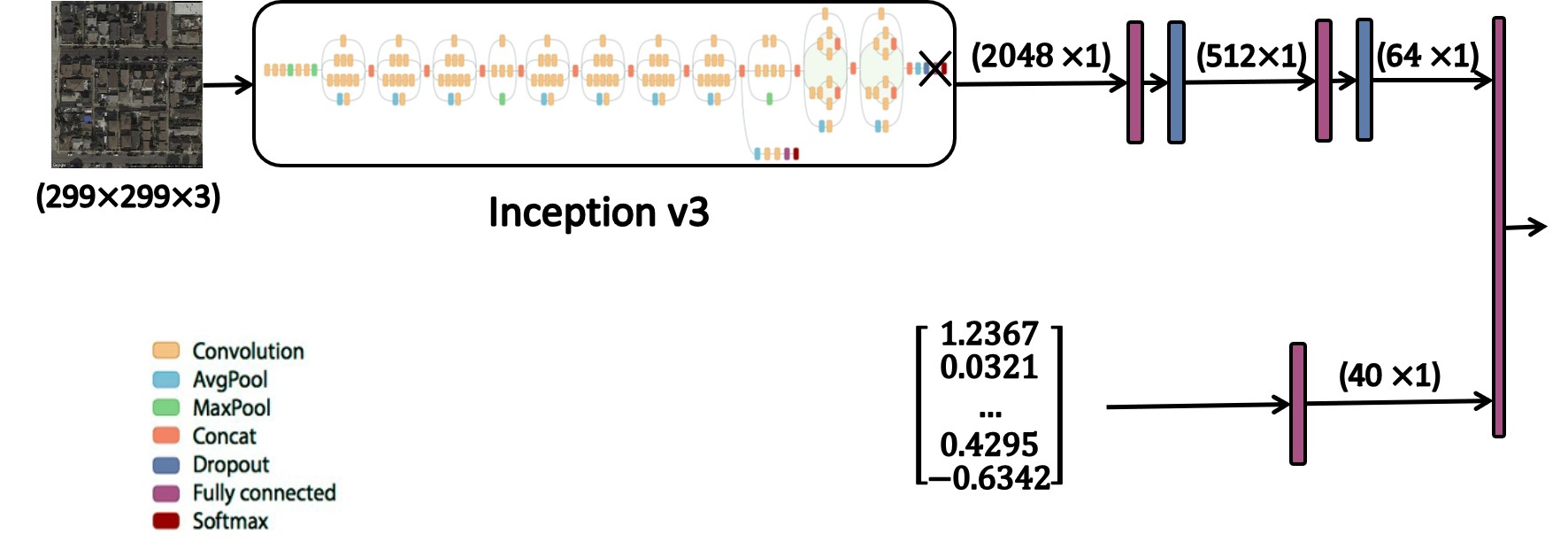}
\centering
\caption{F+I neural network architecture}
\label{fig:architecture}
\end{figure}

We have tried many different architectures and hyperparameters, and in this section only describe the best results. Also, due to lack of space we only describe our model for F+I model.
We use the architecture described in figure \ref{fig:architecture}. The hyper parameters are mentioned in table \ref{table:hyperparameters}. Our model, especially on image side is very prone to over-fitting, so we use L2 regularization and dropout to prevent that. The learning rate (lr) is decreased after each batch is processed, using the following formula:
$lr = \frac{lr}{1 + \alpha \times t}$. We use a relatively big batch size because we were training on a GPU\footnote{NVIDIA Tesla K80}. Larger batch sizes (but smaller than available memory size) help increase the utilization of GPU and generally speed up the training process.

\begin{table}{}
\centering
\begin{tabular}{|c|c|}
\hline
L2 Regularization Parameter & $0.1$\\
\hline
First Dropout (drop probability)&$0.3$ \\
\hline
First Dropout (drop probability)&$0.2$ \\
\hline
Activation Function  & ReLU\\
\hline
Batch Size& $1024$ \\
\hline
Optimizer& Adam(learning rate=$0.0005$, $\beta_1=0.9$, $\beta_2=0.999$)\\
\hline
Epochs& $200$\\
\hline
Learning Rate Decay &$\alpha=0.0001$ \\
\hline
\end{tabular}
 \vspace{0.6em}
\caption{Hyperparameters for F+I model}
\label{table:hyperparameters}
\end{table}

\section{Conclusion/Future Work}
To sum up, by adding satellite images to our prediction model, we were able to improve $R^2$ by ~10\% and reduce MSE by ~60\% compared to the baseline. The results are promising. We believe the following future steps are worth further analysis.

 Using different image zoom levels or a combination of them: Lower zoom levels will be able to capture more details about the neighborhood as they cover a wider range around the property, while a higher zoom level may estimate based on more details from the house, for example the materials used to build it.
 Fine tuning last layers of  the Inception-v3 network: Inception-v3 encodes images of size 299x299x3 to 2048-dimensional vectors. Thus, valuable information may be omitted by this process. As a future step, we can take outputs of earlier layers of the pretrained model or fine tune parameters of the last few convolutional layers. This will allow us to capture more relevant features from the images.
 Adding data from different locations to the dataset to test how generalizable the model is from one location to another: Since public data of property transactions are widely available today, analysis can continue on larger datasets consisting of more counties. This will add to inputs' variance, and potentially make the model more robust. We can then examine if our approach can be generalized.


\medskip

\bibliographystyle{abbrvnat}
\bibliography{references}

\end{document}